\documentclass[preprint,1p,times]{elsarticle}

\usepackage{times}
\usepackage{graphicx}
\usepackage{footnote}
\usepackage{array}
\usepackage{comment}
\usepackage{graphicx}
\usepackage{placeins}
\usepackage{float}
\usepackage{subfigure}
\usepackage[table,xcdraw]{xcolor}
\usepackage{url}
\usepackage{graphicx}
\usepackage{array}
\usepackage[inline]{enumitem}
\usepackage[table]{xcolor}
\usepackage{tabularx}
\usepackage{caption}
\usepackage{float}
\usepackage{tabu}
\usepackage{multicol}
\usepackage{multirow}
\usepackage{longtable}
\usepackage{flushend}
\usepackage{adjustbox}
\usepackage{anyfontsize}
\usepackage{footnote}
\makesavenoteenv{tabular}
\makesavenoteenv{table}
\usepackage{lipsum}
\usepackage{courier}
\usepackage{adjustbox}
\usepackage{booktabs} 

\colorlet{tableheadcolor}{gray!25} 
\colorlet{tablerowcolor}{gray!10} 
\usepackage{booktabs}
\usepackage{colortbl}
\usepackage{amsmath}
\usepackage{xcolor}
\usepackage{graphicx}
\colorlet{tableheadcolor}{gray!25} 
\colorlet{tablerowcolor}{gray!10} 
\usepackage{amssymb}
\usepackage{amsmath}
\usepackage{amssymb} 
\usepackage{hyperref}
\usepackage{url}
\usepackage{csquotes}
\newcommand{\dq}[1]{\enquote{#1}}

\journal{Data Technologies and Applications}

\begin{document}

\begin{frontmatter}



\title{A Data-driven Neural Network Architecture for Sentiment Analysis}

\author{Erion \c Cano\fnref{cor1}} 
%
%
\author{Maurizio Morisio}

\fntext[cor1]{Corresponding Author}

\address{Department of Control and Computer Engineering, Politecnico di Torino, Corso 
Duca degli Abruzzi, 24 - 10129 Torino}





\begin{abstract}
The fabulous results of convolution neural networks in image-related tasks, attracted attention of text mining, sentiment analysis and other text analysis researchers. It is however difficult to find enough data for feeding such networks, optimize their parameters, and make the right design choices when constructing network architectures. In this paper we present the creation steps of two big datasets of song emotions. We also explore usage of convolution and max-pooling neural layers on song lyrics, product and movie review text datasets. Three variants of a simple and flexible neural network architecture are also compared. Our intention was to spot any important patterns that can serve as guidelines for parameter optimization of similar models. We also wanted to identify architecture design choices which lead to high performing sentiment analysis models. To this end, we conducted a series of experiments with neural architectures of various configurations. Our results indicate that parallel convolutions of filter lengths up to three are usually enough for capturing relevant text features. Also, max-pooling region size should be adapted to the length of text documents for producing the best feature maps. Top results we got are obtained with feature maps of lengths 6 to 18. An improvement on future neural network models for sentiment analysis, could be generating sentiment polarity prediction of documents using aggregation of predictions on smaller excerpt of the entire text. 
\end{abstract}


\end{frontmatter}


\section{Introduction} 
\label{intro}
Neural networks are providing ground-breaking results in many complex tasks such as object detection, speech recognition or sentiment analysis. This success is usually attributed to the ability of deep neural networks to generalize well when trained with high quantities of data. 
Among the various neural network types, Convolution Neural Networks (CNN) have become particularly popular because of their ability to mimic the functionality of the human brain visual cortex. As a results, CNNs are applied in image-related tasks such as object detection, fingerprint recognition, computer vision etc. The basic structure of a CNN was first applied by LeCun \emph{et al.} in \cite{lecun1998gradient} for recognizing images of handwritten digits. A decade of hibernation (known as the second AI winter) passed and they showed back in the late 2000s rebranded as \emph{Deep Learning}. At this time they also became essential part of various proposed architectures such as \emph{Inception} in \cite{DBLP:journals/corr/SzegedyIV16}, \emph{ResNet} in \cite{DBLP:journals/corr/HeZRS15} and more.  
Many natural language processing researchers explored use of CNNs or Recurrent Neural Networks (RNN) for text mining tasks such as sentiment analysis, reporting excellent results with little computation load. However, neural network models are usually data hungry and require bigger datasets of training samples. The other problem is the difficulty in finding the optimal hyperparameter setup or design choices when using various types of networks. Optimal network configuration depends on characteristics of available data which should be taken into account. 
\par 
We present in this paper the work we conducted for constructing two relatively big datasets of emotionally labeled songs and the results of many experiments with text datasets of different size and document lengths for simplifying neural network construction. For emotional labeling of songs, we utilized social tags crawled from \emph{Last.fm} music portal. We also adapted a model of music emotions that is highly compatible with the popular model of Russell, together with an annotation scheme based on emotion tags each song has received. Furthermore, the works in \cite{Cano:2018:DLA:3220228.3220229} and \cite{10.1007/978-3-319-77703-0_34} are extended both quantitatively and qualitatively. The first introduces three variants of a neural network architecture that uses convolution and max-pooling layers for text feature extraction and selection as well as a regularized feed-forward layer for classification. In the second paper, various relations between data properties and neural network parameters with respect to optimal performance are explored. In this work we report accuracy scores of a higher number of experiments with more datasets (e.g., including short sentences). Our results can help researchers to simplify hyper-parameter optimization of neural networks that are used for sentiment analysis experiments. 
\par 
%
A fact that we noticed is that bigger datasets are better interpreted by repeating several stacks of parallel convolutions followed by max-pooling layers. An interesting regularity is the one that relates length of documents with pooling region size. The later is the parameter that dictates the size of produced feature maps. According to our results, top scores are achieved when pooling region size is set to produce feature maps that are 6 to 18 units long. Also, convolutions with filter lengths one, two and three are usually enough. Utilizing convolutions of longer filters did not improve results. Regarding the three neural network design we proposed, the basic version with max-pooling layers directly following each convolution layer resulted the best one. The flexibility it offers and its low training time make it a good option as a prototyping basis for practitioners. 
\par 
The rest of the paper is structured as follows: Section \ref{sec:background} presents an overview of various neural network models recently used in text mining tasks. Section~\ref{sec:TagDatasets} describes the steps that were followed for the construction of the two music emotion datasets. Section~\ref{sect:dataParamOpt} presentes preprocessing steps, utilized datasets, and obtained network parameter optimization results. In Section~\ref{sec:alternativeArchs}, we describe the three network architectures we propose. Section \ref{sec:ParamsBaselines} presents the high-level architectural parameters and decisions, together with the literature baselines we compare against. Section~\ref{sec:resultsDiscussion} discusses obtained results and finally, Secton~\ref{sec:conclusions} concludes. 
\section{Background}
\label{sec:background}
\begin{figure}
	\centering
	\includegraphics[width=0.42\textwidth]{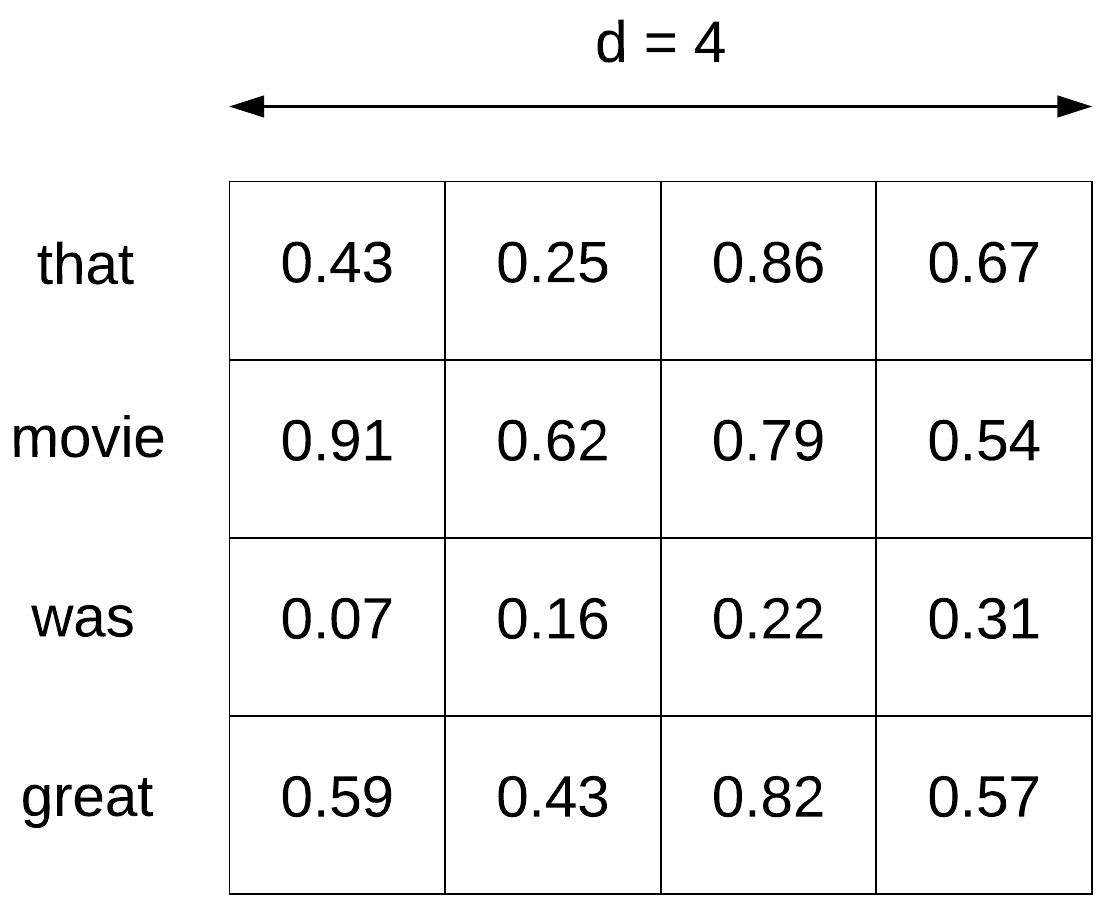}
	\caption{Matrix representation of a movie review}
	\label{fig:TextMatrix}
\end{figure}

Distristributed representations of words known as word embeddings are becoming widely used in text analysis tasks. Their reduced dimensionality makes them well-suited for integration with neural networks. Bag-of-words (BOW) on the other hand, creates a high dimensional space where every word appearing in documents is treated as a feature dimension. Suppose for example that we have a vector of $d$ dimensions for every word appearing in our documents. We can thus represent each document (set of words) as a vector of word vectors (matrix). This is very similar to the matrix of pixels which represents an image. In case we have $d = 4$ (in practice 100 -- 400 dimensions are used), the representation of movie review \dq{that movie was great} will be as shown in Figure~\ref{fig:TextMatrix}. 
%
Because of this similarity, same image recognition techniques like convolution neural networks can be successfully applied on text analysis as well. This started in  late 2000s, with pioneering work conducted by Collobert and Weston in \cite{Collobert:2008:UAN:1390156.1390177} and \cite{Collobert:2011:NLP:1953048.2078186}, applying them on tasks like part-of-speech tagging or named entity recognition. Kim in \cite{DBLP:journals/corr/Kim14f} was one of the first to apply CNNs for sentiment polarity prediction of texts. He used a basic CNN to extract and select features from sentences, reporting excellent performance with little computation load. 
\par 
Many studies that followed explored deeper neural networks of convolution layers. In \cite{Zhang:2015:CCN:2969239.2969312} and \cite{DBLP:conf/eacl/SchwenkBCL17} for example, authors used 9 -- 29 layers, building text pattern features from the basic characters. They reported excellent results on large text datasets of millions of documents. On small datasets however, the deep networks they built are weaker than shallow networks which start from word representations. 
A discussion and empirical comparison of shallow word-level convolutional networks against deep character-level ones can be found in \cite{DBLP:journals/corr/Johnson016a}. They applied shallow networks on the same big datasets that were created and used in \cite{DBLP:conf/eacl/SchwenkBCL17} and \cite{Zhang:2015:CCN:2969239.2969312}. According to their results, word-level networks are both more accurate and also considerably faster than deep character-level ones. A possible explanation is the semantic value contained in language words, which is completely lost when analysis starts from characters. On the other hand, a handicap of word-level CNNs is the high number of parameters they need for representing text words and thus higher memory requirements during training. 
\par 
Yet another family of neural networks that has gained considerable popularity in the recent years, is that of RNNs which work particularly well on data which exhibit time continuity like word sequences, time series etc. The most popular implementation of RNNs is probably the Long Short-Term Memory (LSTM) network presented in \cite{Hochreiter:1997:LSM:1246443.1246450}. Its applicability on sentiment analysis research has taken off recently with works like \cite{Lai:2015:RCN:2886521.2886636}. There are also several studies like \cite{conf/coling/ZhouQZXBX16} that combine them with CNNs for better results. 
Research in text mining is apparently following the same path as that in image recognition where complex architectures of many layers have been introduced. Same as with images, best results on texts are achieved with models that are complex in the feature extraction and selection part, but shallow and quiet simple in classification. 
In \cite{DBLP:conf/acl/JohnsonZ17} for example, we find a deep architecture of many convolution layers, followed by max-pooling layers placed after certain stacks of consecutive convolutions. Few feed-forward layers are used at the end of the network for classification. Authors try their architecture on big datasets created in \cite{Zhang:2015:CCN:2969239.2969312} and report competitive results.     
\par 
Also in \cite{DBLP:conf/coling/SantosG14}, a similar yet simpler architecture puts together information gained from characters, words and sentences. Each character appearing in training texts is encoded in fixed-length character vectors, creating an embedding matrix. Character embeddings are used to capture useful shape or morphological patterns such as hash tags inside words. Word-level embeddings are also generated using the popular Skip-Gram method applied on a collection of 1.75 billion tokens from English Wikipedia. What follows is a convolution layer that extracts local features and max-pooling for creating the feature vectors of the entire sentence. The derived model is evaluated on Stanford Sentiment Treebank dataset of sentences and Stanford Twitter Sentiment collection of tweets, reaching peak results on both.  
\par 
\section{Creating Song Datasets from Social Tags}   
\label{sec:TagDatasets}
The need to experiment with data-driven predictive models has motivated researchers to utilize various strategies for creating datasets of labeled training samples. However, due to copyright enforcements on data that are shared online, it is hard to find free and open datasets to experiment with. A problematic domain is especially music that is highly commercial and copyrighted.  As a result, researchers who work in music emotion recognition or music information retrieval do usually face difficulties due to the lack of data. In this section, we present the creation steps of two relatively big song emotion datasets utilizing social tags of \emph{Last.fm} music portal. 
\subsection{Folksonomy of Song Emotions}
\label{sec:folksonomy}
When experimenting with predictive models for music emotion recognition, collections of emotionally labeled songs and feature extractions tools and/or skills are indispensable. In that scenario, an ideal datasets of songs should possess the following four characteristics:
\begin{enumerate}[label=\textbf{\arabic*}.]   
	\item \emph{Contain as many songs as possible (e.g., more than 1000)}
	\item \emph{Annotated following a well-known model of emotions}
	\item \emph{Have polarized annotations to be usable as ground truth}
	\item \emph{Publicly released for cross-interpretation of experiments}
\end{enumerate}
First, it should be as big as possible to yield good predictive models. Second, emotionality of songs should follow a well-known model of emotions that is popular among researchers. An example could be the dimensional model of Russell that was presented in \cite{russell1980circumplex}. Regarding the third requirement, contained songs should clearly belong to the emotional category they are assigned with. Finally, the dataset should be open and freely available for cross-interpretation of experimental results by different researchers. 
When searching for existing music emotion datasts we observed that the above characteristics are conflictuing and hardly achieved together. There were several open datasets but they were mostly small.
The one described in \cite{7536113}, for example, was created using \emph{AllMusic} tags and follows the popular valence-arousal model of Russell. However, it is relatively small, consisting of 771 tracks only. The big datasets on the other hand were restricted from public access. Such example is the dataset of 5296 songs described in \cite{hu2009lyric}.
Given that we did not find any music dataset complying with the four requisites listed above, we decided to create one ourself. The basic source of intelligence we could access were social tags of \emph{Last.fm} users. They are terms that describe different aspects of songs such as genre, emotions, instrumentation, etc. 
To comply with the second requisite, we explored various literature folksonomies derived from social tags. The one of \cite{conf/ismir/BischoffFPNLS09} for example, was derived by clustering \emph{AllMusic} mood tags and aggregate them in four categories that are analogous to the four quadrants of Russell's model. The work in \cite{conf/ismir/LaurierSSH09} utilized \emph{Last.fm} emotion tags to derive a very similar folksonomy of four music emotion classes (\emph{happy}, \emph{angry}, \emph{sad}, \emph{relaxed}) that are again analogous to the four quadrants of Russell. 
All these research results convinced us that among the the categorical version of Russell's model with one emotional category for each quadrant (\emph{happy} for Q1, \emph{angry} for Q2, \emph{sad} for Q3 and \emph{relaxed} for Q4) is the most simple, popular and practical model for our purpose. 
\begin{table}
	\caption{Four clusters of tag terms}
	\centering
	\begin{tabular}{l l l l}   
		\toprule	
		\multicolumn{1}{l}{\textbf{Q1-Happy}} & \multicolumn{1}{l}{\textbf{Q2-Angry}} &
		\multicolumn{1}{l}{\textbf{Q3-Sad}} & \multicolumn{1}{l}{\textbf{Q4-Relaxed}}  \\
		\midrule
		happy & angry & sad & relaxed \\
		happiness & aggressive & bittersweet & tender\\
		bright & fierce & bitter & soothing \\
		joyous & outrageous & sadness & mellow \\	
		cheerful & rebellious & depressing & gentle \\	
		fun & anxious & tragic & peaceful \\
		humorous & fiery & gloomy & soft \\
		merry & tense & miserable & calm  \\
		exciting & anger & funeral & quiet \\
		silly & hostile & sorrow & delicate \\
		\bottomrule
	\end{tabular}
	\label{table:ClassModel}
\end{table}
\par
\begin{figure}
	\centering
	\includegraphics[width=0.6\textwidth]{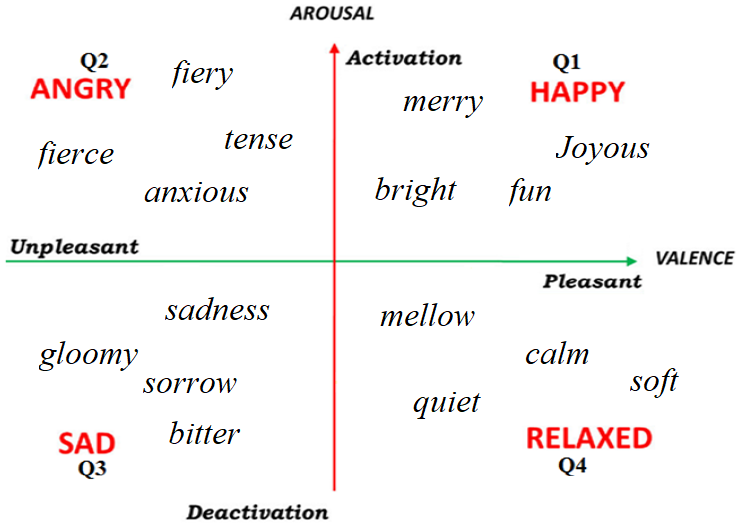}
	\caption{Adopted model of tag emotion categories}
	\label{fig:MyTagFolk}
\end{figure}

 
For our purpose, we constructed a folksonomy that is very similar to the one of \cite{conf/ismir/LaurierSSH09}. First, we retrieved more than 100 emotion terms from relevant research papers and extended them with synonyms. An initial selection was conducted manually to filter out ambiguous terms and keep in only those that clearly fall in one of the four categories. For objectivity in selection, we consulted ANEW valence and arousal norms of each word. To check the correctness of our processing and the quality of the folksonomy we measured the average intra-cluster (as high as possible) and inter-cluster (as low as possible) similarities. For the computations we utilizedword embeddings trained from a corpus of 2 billion tweets with Glove method.\footnote{\url{http://nlp.stanford.edu/data/glove.twitter.27B.zip}} Word embeddings are very popular for capturing semantic similarities between words. The average intra-cluster similarities were optimized by performing an exhaustive combinatorial search among the preliminary tags in each of the four clusters. The optimal configuration resulted the one shown in Table~\ref{table:ClassModel} with  ten most suitable mood tags in each cluster. A graphical illustration of the model is shown in Figure~\ref{fig:MyTagFolk}. 
Further details about music emotion models and the steps we followed to derive our folksonomy can be found in \cite{phdthesis} (Chapter 3).  
\subsection{Data Processing and Annotation}
To produce a large set of songs with polarized emotional labels (first and third requisites) we imported all tracks of \emph{Million Song Dataset} described in \cite{Bertin-mahieux11themillion}. Other collections of newer songs such as Playlist dataset were also mixed in. After removing duplicates we reached to a initial collection of 1018596 tracks. At this point, all tags of each track were crawled utilizing \emph{Last.fm API}.\footnote{\url{https://www.last.fm/api/show/track.getTags}} We moved on removing every tag that was not about mood and then songs that were left with no tags. At the end of this phase, we reached to 288708 tracks.We saw that there was a bias towards positive emotion tags. This is probably because people are more inclined to give feedback when they listen to positive songs. 
Further details about data processing steps and tag statistics can be found in \cite{pub2669975}.     
Next, emotion tags of each cluster appearing in the remaining songs were identified and counted. To comply with the third requirement we used a tight scheme for the annotations. 
A song is set to cluster $Qx$ if it fulfills one of the following conditions: 
\begin{itemize}
	\item has 4 or more tags of $Qx$ and no tags of any other quadrant
	\item has 6 up to 8 tags of $Qx$ and at most 1 tag of any other quadrant
	\item has 9 up to 13 tags of $Qx$ and at most 2 tags of any other quadrant
	\item has 14 or more tags of $Qx$ and at most 3 tags of any other quadrant
\end{itemize}
The above scheme ensures that even in the worst case, any song set to Qx cluster has more than 75\% of all its received tags being part of that cluster. The output was a set of 1986 happy or $Q1$, 574 angry or $Q2$, 783 sad or $Q3$ and 1732 relaxed or $Q4$ songs for a total of 5075 (2,000 after balancing). We also create a datasets with \emph{Positive vs. Negative} representation by merging $Q1$ with $Q4$ (\emph{happy} with \emph{relaxed}) considering them as \emph{positive}, and $Q2$ with $Q3$ (\emph{angry} with \emph{sad}) for the \emph{negative} category. The corresponding tags of each cluster were recombined accordingly. As binary discrimination is easier, we used an even tighter annotation scheme. A song was considered to belong to Qx (\emph{positive} or \emph{negative}) only if:  
\begin{itemize}
	\item it has 5 or more tags of $Qx$ and no tags of the other category
	\item it has 8 up to 11 tags of $Qx$ and at most 1 tag of the other category
	\item has 12 up to 16 tags of $Qx$ and at most 2 tags of the other category
	\item has 16 or more tags of $Qx$ and at most 3 tags of the other category
\end{itemize}
This scheme guarantees that even in the worst case, any song labeled as \emph{positive} or \emph{negative} has more than 85\% of all its received tags being part of that category. The output was a collection of 2589 \emph{negative} and 5940 \emph{positive} songs, for a total of 8529 (5,000 after balancing). 
Both datasets can be freely downloaded from our group website.\footnote{\url{http://softeng.polito.it/erion/}} Lyrics, or metadata of their songs can be easily retrieved from online music websites.   
\section{Data-driven Optimization of Network Parameters}
\label{sect:dataParamOpt}
It is not easy to apply convolution or other types of neural networks for sentiment analysis or similar tasks. The simplest network architecture could be a convolution layer over word embeddings for feature extraction, followed by a max-pooling layer for feature selection. In this case we have a fixed network depth $D = 2$, but there are still many other parameter values to pick up. We have to decide about number of convolution filters $m$, filter length $k$, filter stride $s$, pooling region length $r$, as well as regularization norms like dropout, $L_1, L_2$ etc. It is hard to find an optimal setup of all these parameters. On the other hand, model performance is highly influenced by input data characteristics such as training set size, length of documents etc. As a results, it makes sense to try various input data and observe patterns that relate their characteristics with network settings with respect to optimal network classification accuracy. In this section we conduct a series of experiments that try to answer the following questions: 
\begin{description}
	\item[RQ1] \emph{How is document length related with pool region size ($r$)  
		with respect to optimal accuracy score?}
	\item[RQ2] \emph{What effect do convolution kernel size ($k$) and network width
		($W$) have on prediction accuracy?} 
	\item[RQ3] \emph{How is training dataset size related with network depth ($D$) 
		with respect to optimal accuracy score?}
\end{description}
The following subsections describe the preprocessing steps, experimental settings, obtained results and concluding remarks addressing the above three questions. 
\subsection{Experimental Setup}
A first important decition to make is how to create or find word embedding representations for each text document. One possibility is to initialize vectors for each word with random values and later use the neural network for tuning them during training. This method requires a large training collection of documents. As \cite{pub2668229} suggests, making use  of pretrained word vectors created from big text corpora gives higher performance when small training datasets are available. Considering that the datasets we use here are small to medium-sized, we decided to source word vectors from a pretrained collection.\footnote{\url{https://code.google.com/p/word2vec/}} It contains 300-dimensional vectors of words appearing in a corpus of 100 billion GoogleNews tokens. Same vectors were successfully utilized in various relevant studies like \cite{DBLP:journals/corr/Kim14f} or \cite{DBLP:journals/corr/LauB16}.
\par
\begin{figure}
	\centering
	\includegraphics[width=0.58\textwidth]{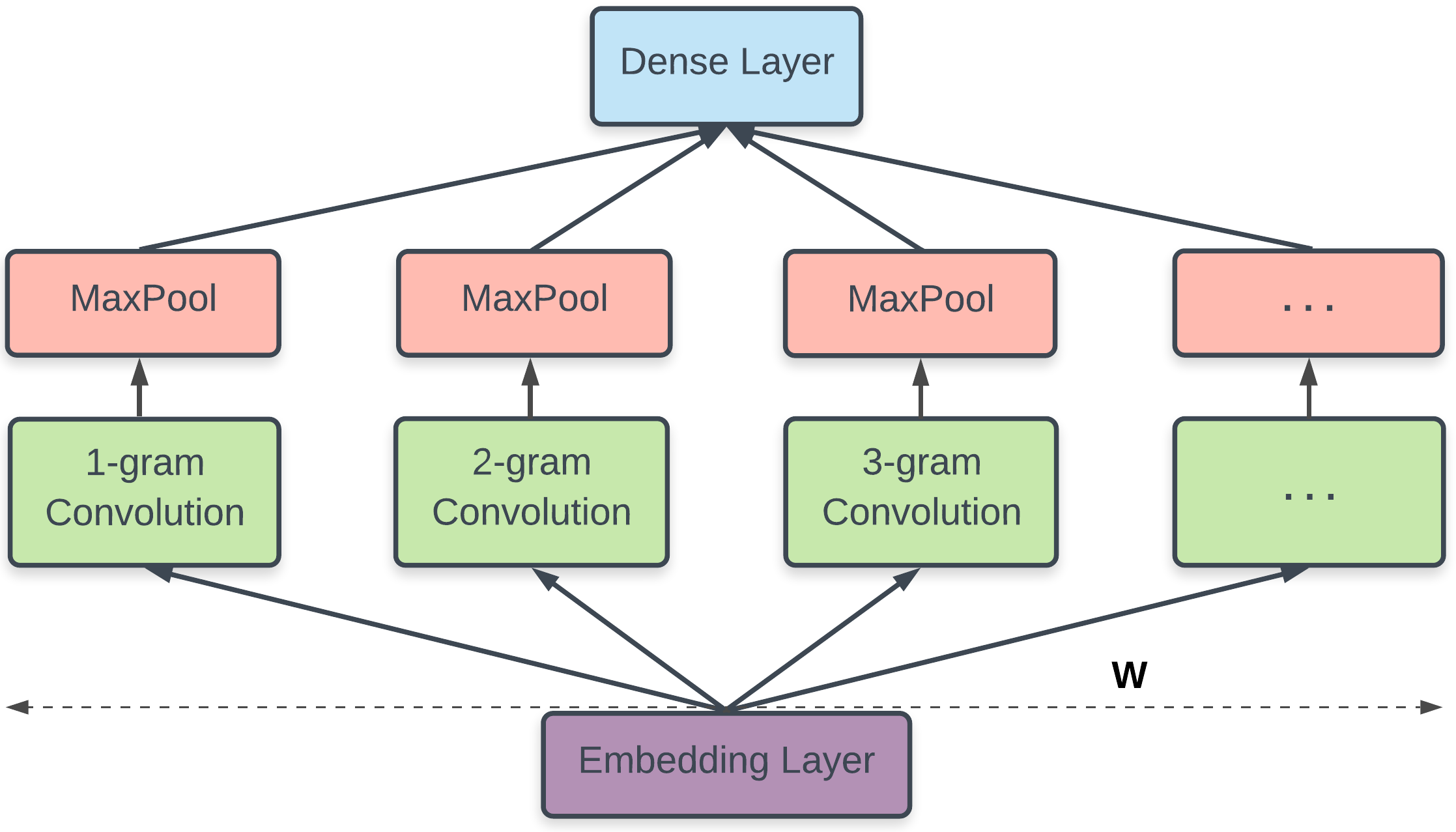}
	\caption{Basic neural network structure}
	\label{fig:BasicNetStruct}
\end{figure}

 
The network model chosen for the experiments is schematically presented in Figure~\ref{fig:BasicNetStruct}. First comes the embedding layer which is not trainable. Pretrained vectors for each word appearing in documents are directly sourced. Next we have a stack of convolution layers that are used for feature extraction. There are $W$ of them with filter sizes $k = 1$, $k = 2$, $k = 3$ and so on. All these layers are used in parallel for extracting word, 2-gram, 3-gram and longer features. 
A fixed number of 80 filters is applied in each layer to produce the feature maps.
%
An important parameter of the network is its depth ($D$) or the total number of layer stacks. This experimental network of Figure \ref{fig:BasicNetStruct} has a depth $D = 2$, with a stack of max-pooling layers follows the convolutions. Its role is to subsample data as well as to select the most relevant and fruitful features for the classification phase. We utilize regional max-pooling which is renowned for its ability to retain some local information of word combinations in texts. Furthermore, relevant works such as \cite{DBLP:conf/eacl/SchwenkBCL17} and \cite{DBLP:journals/corr/ZhangW15b} confirm that it performs better than other pooling schemes like average or k-max. As for length of region $r$, we experiment with different values on the different datasets. Feature maps that are produced have length $l = ceil(\frac{n - k + 1}{r})$. Here $n$ is the number of words in the training documents. The feature maps are concatenated and pushed to a feed-forward layer that works as classifier. 
The latter is made up of 80 nodes. A 0.1 $L_2$ regularization norm and 0.35 dropout rate were used to mitigate overfitting. Binary crossentropy and Adam optimizer were used to compute loss and optimize training. 
During the experiments, a 70/10/20 \% data split for training, development and testing respectively was used. 
\begin{figure}
	\centering
	\includegraphics[width=0.58\textwidth]{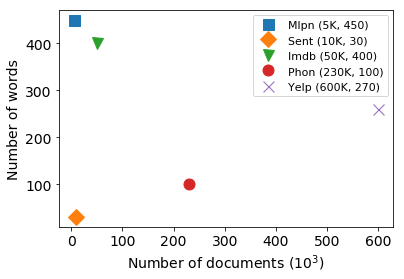}
	\caption{Size-length distribution of training datasets}
	\label{fig:size-words-plan}
\end{figure} 
To find optimal network structure and observe combinations of data properties and network parameters associated with it, we explored alternative versions derived from that of Figure \ref{fig:BasicNetStruct}. Decisions about parameters of these alternative networks were driven by properties of training data. For example, various values of max-pooling region length $r$ are used to adapt to short, medium or long-sized text documents. Furthermore, deeper network structures ($D > 2$) are used in conjunction with the bigger training datasets. Those structures are obtained by simply duplicating the convolution and pooling stacks of the basic network.   
\subsection{Datasets and Statistics}
\label{subsec:datasets}
Public datasets of different sizes and text lengths were chosen for the experiments. They contain text units of various thematic contents such as movie reviews, smart phone reviews or song lyrics and are described below. 
\begin{description} 
	\item [Yelp] This is the largest dataset we found and used. It was produced by 	from Zhang \emph{et al.} in \cite{DBLP:journals/corr/ZhangZL15} and consists of 598,000 Yelp user reviews about places such as hotels, restaurants, etc. Each text review was given the corresponding emotional polarity label. 
	\item [Phon] Here we have a collection of user reviews about smart phones sold in Amazon. Besides the text comments, users often provide 1--5 star rating also. The dataset has records of both star rating and text comments. Records with 3-star rating were removed to have a polarized separation of 1-star and 2-star reviews considered as \emph{negative}, from 4-star and 5-star reviews that were denoted as \emph{positive}. A total of  232546 reviews was finally obtained.  
	\item [Imdb] This dataset is the bigger and more recent IMDB collection of movie reviews presented in \cite{Maas:2011:LWV:2002472.2002491}. It comprises comments about 50,000 movie, manually annotated as \emph{positive} or \emph{negative}. Being bigger in size, it been used as experimental basis for more recent studies involving neural networks. IMDB movie review dataset is also available in various libraries,  in a preprocessed and ready to use format. 
	\item [Sent] Sentence polarity dataset is a collection of \emph{positive} and \emph{negative} 	sentences that has been used since 2005 in many other studies.  It was created by Pang and Lee in \cite{Pang:2005:SSE:1219840.1219855}. The collection contains 5331 \emph{positive} and 5331 \emph{negative} texts originating from IMDB movie archive. Each sentence is short, consisting of 10 -- 20 words.   
	\item [Mlpn] MoodyLyrics is a collection of emotionally-labeled songs presented in \cite{ismsi17}. To conform with the other binary (\emph{positive} vs \emph{positive}) tasks, we use MoodylyricsPN, a derivative collection of  2,500 \emph{positive} and 2,500 \emph{negative} songs described in \cite{pub2669975}. The corresponding lyrics were scrapped from Web music portals and are used for experimental purposes only.  
\end{description}
Today it is certainly useful to have large supervised datasets for experimentation. However, annotation that is based on simple heuristics like star rating only is probably not enough and further data quality evaluations are probably required.    
\begin{table}[ht] 
	\caption{Document length statistics for each dataset}  
	\small 
	\centering    
	\setlength\tabcolsep{2.7pt}  
	\begin{tabular}
		{l c c c c c}  
		\toprule
		\textbf{Dataset} & \textbf{\#Docs} & \textbf{MinLen} & \textbf{AvgLen} & 
		\textbf{MaxLen} & \textbf{UsedLen}		 	\\ 
		\midrule   
		Yelp Reviews & 598K & 1 & 122 & 963	& 270	  \\	
		Phone Reviews & 232K & 3 & 47 & 4607 & 100	  \\ 
		Movie  Reviews & 50K & 5 & 204 & 2174 & 400		  \\
		Sentence Polarity & 10K & 1 & 17 & 46 & 30	  \\ 
		Song Lyrics & 5K & 23 & 227 & 2733 & 450	  \\
		\bottomrule
	\end{tabular} 
	\label{table:MP8documentStatistics}
\end{table}
\par 
Each document of the five datasets was preprocessed following common procedures. As many of the texts were scrapped from websites, any remaining html tags were first removed. Contrary, smiley symbols like :(, :-(, :P, :D, :-), :) were keept in. They are highly related with text emotionality and thus represent very relevant features. Stopwords are removed in most of studies, as they usually contain very little or no semantic value. Here we discarded the subset \{\dq{for}, \dq{an}, \dq{as}, \dq{by}, \dq{the}, \dq{these}, \dq{those}, \dq{this}, \dq{of}, \dq{at}, \dq{that}, \dq{a}\}. Besides smiley symbols, short for residues like \dq{s}, \dq{t}, \dq{m}, \dq{ll}, \dq{d} and negations such as \dq{couldn}, \dq{don}, \dq{hadn} or \dq{didn} were also kept in. Their presence or absence can shift the emotional polarity of phrases, sentences or entire document. The remaining \dq{errornous} patterns were also removed and everything was finally lowercased.  
\par 
For the conducted experiments, it was important to have an idea about document lengths. Length statistics of texts in each dataset are listed in Table~\ref{table:MP8documentStatistics}. As we can see, Yelp review lengths span between 1 and 963 words, averaging at 122. Smart phone and movie review lengths are more dispersed. The former range from 3 to 4607 with average of 47 words. The later fall between 5 to 2174 averaging at 204. Sentence polarity dataset has the shortest and most uniform texts. They range between 1 and 46 words, averaging at 17. Song lyrics on the other hand are the longest, with length span from 23 to 2733 and average of 227 words. Length distribution statistics revealed that most documents in each dataset are much shorter than the average. In smart phone and Yelp review datasets for example, we found very few datasets longer than 500 words. For this reason and also to reduce computation load of experiments, the few long documents were clipped and the shorter ones were zero-padded to reach a uniform length. The experimental length of each dataset is reported in the last column of Table~\ref{table:MP8documentStatistics}.
Furthermore, in Figure~\ref{fig:size-words-plan} we can see a planar projection of document length and size in each datasets. 
\subsection{Network Optimization Results}
\label{subsec:OptResults}
We performed various experiments using the setup and datasets described above. Top performing neural networks and the corresponding accuracy scores on each dataset are listed in Table~\ref{table:mp8FiveNetworkScores}. In general we see that simple network structures of just one or two convolution and max-pooling stacks yield very good results. Highest score achieved on phone reviews seems excellent. The one on song lyrics on the other hand, is not that good. In fact, we did not find any experimental result on these two datasets to compare with. Peak score on sentence polarity dataset is 79.89\%. The many authors that have used it, report accuracy scores from 76 to 82\%. The record result reported in the literature is actually 83.1 \%. It is reported by Zhao \emph{et al.} in \cite{DBLP:journals/corr/ZhaoLP15} where a self-adaptive sentence model is proposed. Their model uses gating networks to form a hierarchy of representations from words to sentences and finally to entire document.   
Best score obtained on movie reviews dataset is 90.68\%. Top score reported in literature is 92.23 \%. It was reached by Zhang and Johnson in \cite{DBLP:journals/corr/Johnson014}. Same authors also reach a peak score of  97.36 \% on Yelp reviews in \cite{DBLP:conf/acl/JohnsonZ17}. For both results, they utilized deep and highly complex
neural networks. 
\begin{table}[ht] 
	\caption{Accuracies of top five network structures}  
	\centering    
	\setlength\tabcolsep{4.4pt}  
	\begin{tabular}
		{l c c c c c}  
		\toprule
		\textbf{Network} & \textbf{Mlpn} & \textbf{Sent} & \textbf{Imdb} & \textbf{Phon} & \textbf{Yelp}	\\ 
		\midrule 
		3 x Conv-Pool3 & 71.88 & 74.04 & 89.11 & 95.21 & \textbf{94.95}  \\ 		
		2 x Conv-Pool5 & 75.44 & 74.22 & \textbf{90.68} & 95.64 & 93.84  \\ 
		2 x Conv-Pool4 & 73.34 & 75.08 & 89.87 & \textbf{96.57} & 94.86	  \\ 
		Conv-Pool25 & \textbf{75.63} & 74.46 & 90.12 & 95.15 & 93.51	  \\ 	
		Conv-Pool4 & 72.24 & \textbf{79.89} & 87.98 & 95.31	& 92.32	  \\
		\bottomrule
	\end{tabular} 
	\label{table:mp8FiveNetworkScores}
\end{table}                                                                                                                         
\par 
Even though lower that literature records, our results were reached with very simple and quick models of less than 200,000 parameters. Contrary, the top models we compared with are complex neural networks with millions of trainable parameters. 
Digging deeper in Table \ref{table:mp8FiveNetworkScores}, we observe interesting patterns for answering the three questions posed in the beginning of this section. Regarding RQ1, we see that datasets of longer documents (Yelp, Imdb and Mlpn) perform better on models with longer pool regions. Top scores we got on those datasets were reached with aggregate max-pooling coefficients $r = 3~x~3~x~3 = 27$, $r = 5~x~5 = 25$ and $ r = 25$ respectively. Contrary, Phon and Sent datasets that contain shorter texts, reach peak scores on models of smaller region. Actually $r$ is the parameter that dictates size of final feature maps ($l = ceil(\frac{n - k + 1}{r})$). According to the results, highest accuracy scores are always reached with $l$ values that are within $6 - 18$.   
\par 
Regarding filter size effect on performance (RQ2), convolutions with $k = 1, 2 $ and $3$ were essential for optimal scoring. However, networks of more convolutions with filter sizes $k = 4$ or $k = 5$ did not improve performance. Consequently, we used networks of $W = 3$.
Regarding RQ3 and network depth, smallest datasets (Mlpn and Sent) reach their peak scores on the most simple network structure of two stacks. They are overfitted on the more complex networks which are obviously data hungry. Doubling convolution-pooling stacks from $D = 2$ to $D = 4$ improves accuracy on the two bigger Phon and Imdb datasets. Yelp which is the largest dataset, achieves top score on an even deeper network of six convolution-pooling layer stacks. To keep the models simple, we did not try networks of $D > 6$. However, anyone who has tons of labeled data can explore deeper network configurations. Also, more complex network designs can produce even higher results. In the following section we address our efforts precisely on this issue. 
\section{Alternative Convolution Architecture Designs}
\label{sec:alternativeArchs}
\begin{figure}[h!]
	\centering
	\includegraphics[width=0.65\textwidth]{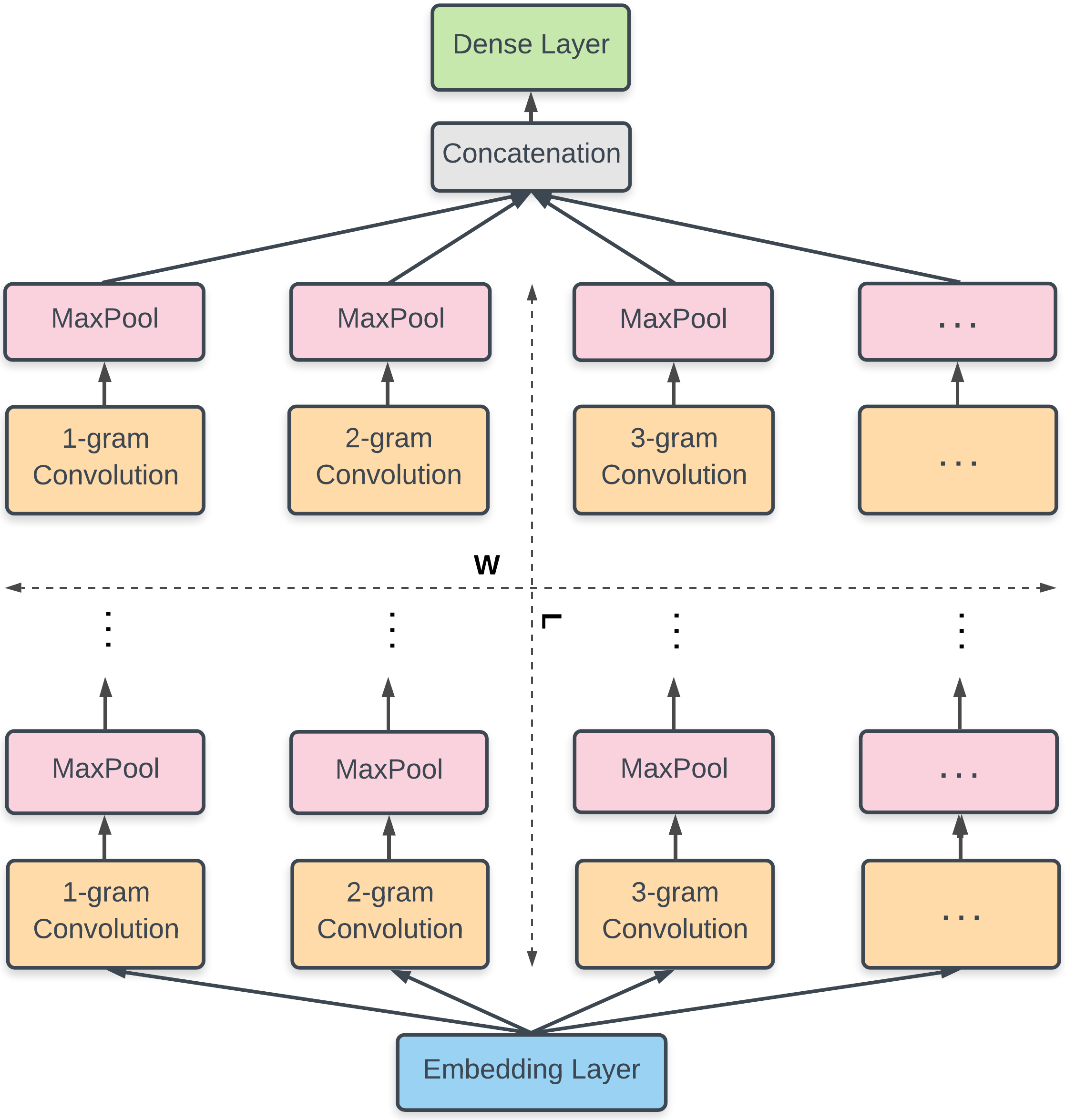}
	\caption{Basic architecture scheme}
	\label{fig:NgramCnn1}   
\end{figure}

Based on the results of the previous section, we performed a better evaluation of the three network architectures we presented in \cite{Cano:2018:DLA:3220228.3220229}. The basic version that is shown in Figure \ref{fig:NgramCnn1} (NgramCNN), is just a vertical extension and generalization of the structure in Figure~\ref{fig:BasicNetStruct}. The first layer creates vector representations of words seen in training documents. After that, a series of $W$ parallel convolutions with increasing filter sizes ($k=1, \ldots, W$) follow. Convolution operations work as word and phrase feature extractors. $W$ (width) can vary in response to the task and the type of data that are being analyzed. A stack of max-pooling layers with region $r$ follows. Their role is to reduce the feature maps by selecting the most relevant features for classification. Same as with $W$, the value of $r$ can be adjusted to adapt to number of words in each document of training set. Same convolution and max-pooling stacks are repeated several times, forming a network of $D$ stacks in total. Once again, the value of $D$ is flexible and should be chosen based on size of training datasets. The output ot the last max-pooling layers is the final set of features that will be used. They are pushed to a fully-connected dense layer for classification.   
\par
\begin{figure}[h!]
	\centering
	\includegraphics[width=0.65\textwidth]{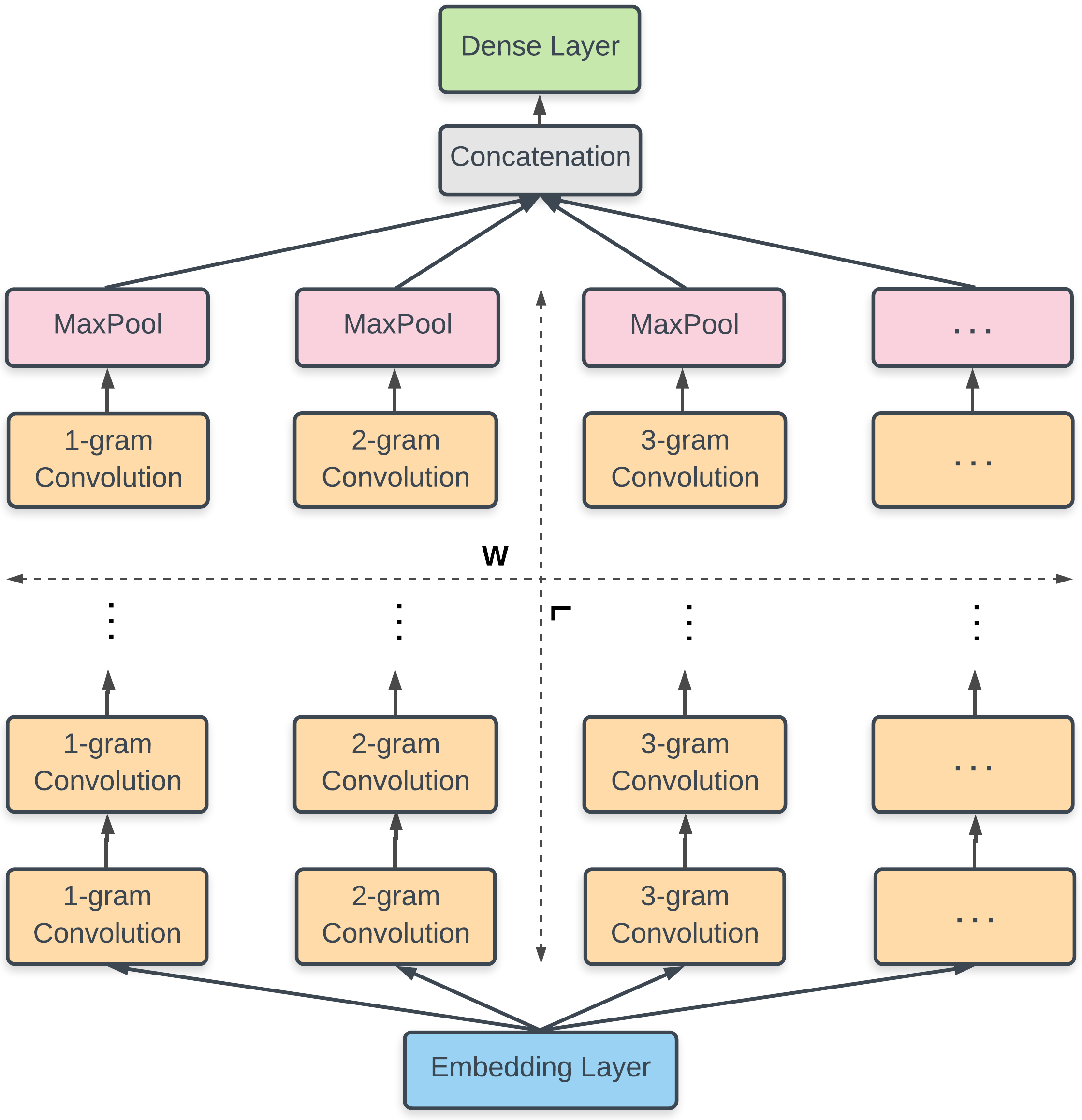}
	\caption{Pyramid architecture with downsampling convolutions}
	\label{fig:NgramCnn2}   
\end{figure}

 
Figure~\ref{fig:NgramCnn2} shows the second variant which is slightly different. Here we have an Embedding layer, a first stack of convolutions, the last stack of max-pooling layers and the classification layer, that are same as in the basic architecture. The difference is in the downsampling layers of intermediate stacks. Instead of regional max-pooling, convolutions with stride $s > 1$ are used instead. If a fixed value of $s$ (e.g., $s = 2$) is utilized in all layers, feature maps are reduced equally in every stack. Consequently, the feature sets become pyramid-shaped and similar to those of the model presented in \cite{DBLP:conf/acl/JohnsonZ17}.
\par 
There are few differences to note when comparing striding convolutions with max-pooling layers. Striding convolutions are able to preserve some positional information in feature sets, which may be an advantage for classification. Simple max-pooling on the other hand, forgets everything about the spacial structure of data features. It only selects the features that seem the most relevant, regardless of their position in the set. One problem with convolutions is their training time. Because they have many trainable parameters, they are slower than max-pooling layers. A more detailed comparison between striding convolutions and pooling can be found in \cite{DBLP:journals/corr/SpringenbergDBR14} where several experiments for object-recognition via convolutions are conducted. Authors suggest that striding convolutions can completely replace max-pooling layers, yielding simpler network architectures without performance loss. Their results are still limited on image analysis only and do not consider other types of tasks such as text mining or sentiment analysis.  
\par
\begin{figure}[h!]
	\centering
	\includegraphics[width=0.65\textwidth]{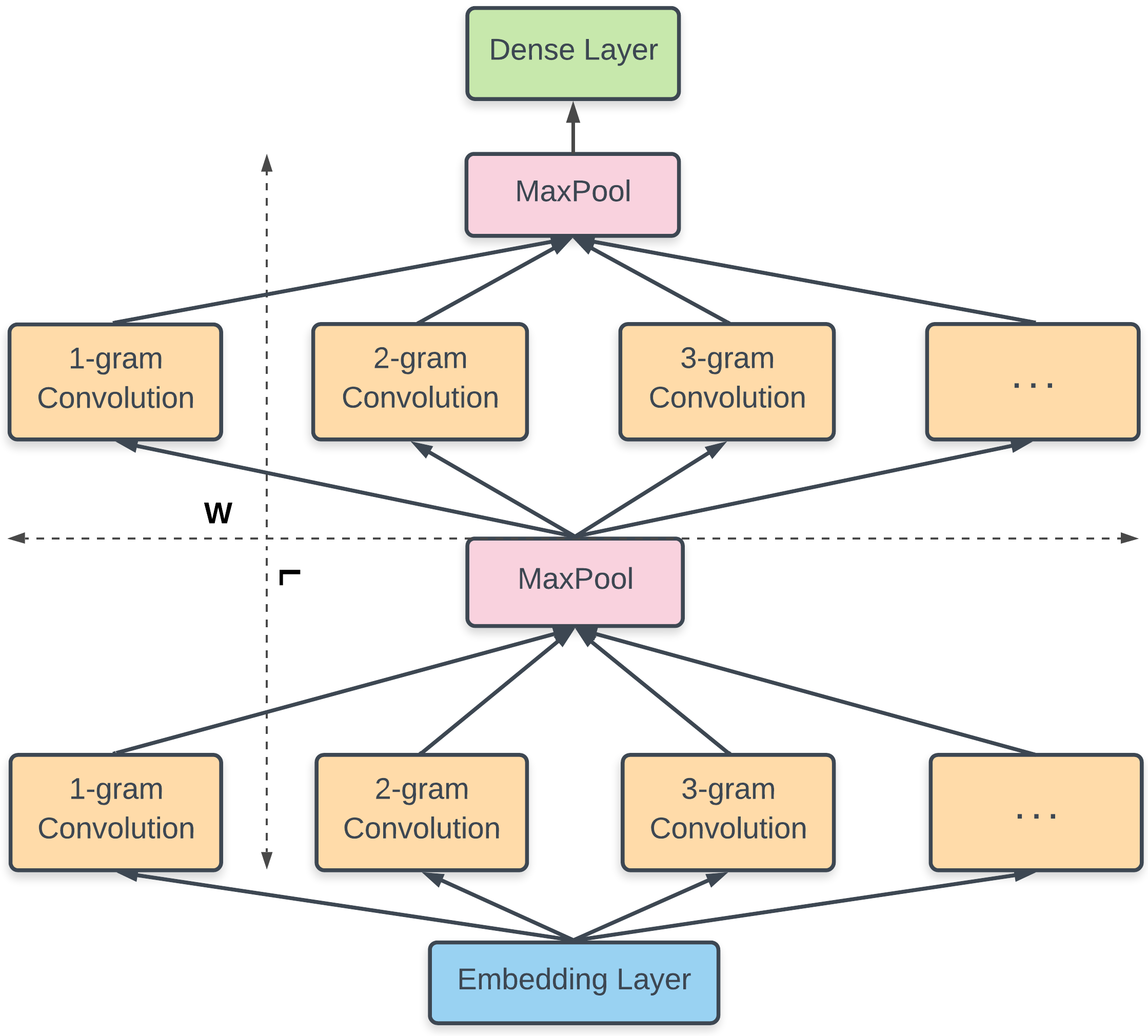}
	\caption{Architecture with fluctuating features}
	\label{fig:NgramCnn3}   
\end{figure}

 
The third architecture version is presented in Figure \ref{fig:NgramCnn3}. It is significantly different from the two others which were pure multi-channel neural network structures. First and last layers are the only similar parts. Feature selection and extraction is organized in a fluctuating form, with feature maps expanding and contracting after each convolution and max-pooling stack.  
The first $W$ layers of convolutions create a wide set of features which are concatenated and pushed to the first max-pooling stack. This later has region size $r$ and significantly reduces (contracts) features' size. Later on, a second turn of convolutions is applied in the features, expanding their size once again. The process is repeated with a second max-pooling stack for subsampling, creating a total number of $D$ network stacks. Same as in the other architectures, $r$, $W$ and $D$ values can be chosen based on properties of available data. 
\section{Hyperparameters and Baselines}
\label{sec:ParamsBaselines}
This section presents several network parameter decisions made for comparing the three network architectures described above. These experiments extend over those presented in Section~\ref{sect:dataParamOpt}. Other neural network models presented in similar text mining studies are also described and used as comparison baselines. 
\subsection{Common Network Settings}
\label{NgCnnSettings}
A common setup of basic parameters was chosen for all three network architectures presented in the previous section. Based on the results of Section~\ref{subsec:OptResults}, we set $W$ to 3, omitting convolutions of longer filters. As for network depth, four stacks of layers were best performers in most previous experiments. For this reason, four stacks ($D = 4$) were chosen for feature extraction and selection. 
We reused four of the five datasets described in Section \ref{subsec:datasets}, excluding Mlpn only. This later dataset is quiet small and was overfitted in many previous experiments. Furthermore, we do not have baseline results from others studies on it.
Region sizes $r$ of max-pooling layers were also chosen based on results of Section~\ref{subsec:OptResults}. Since we use hierarchical pooling, the product of region sized in the two stacks must equal the optimal value that is to be used. For this region, $r = 2, ~ r = 5, ~ r = 4$ and $r = 5$ were chosen for Sent, Imdb, Phon and Yelp datasets respectively. In the case of the pyramid-shaped version, the this values of $r$ correspond to the convolution stride ($s$) of the second stack. 
Regarding text preprocessing, same steps as those of Section~\ref{subsec:datasets} were also applied. For document representations, we once again used pretrained word vectors sourced from GoogleNews.
\subsection{Baseline Models}   
\label{NgramCnnBaselines}
With network settings described above, we compare the three network architectures of Seciton \ref{sec:alternativeArchs} with each other as well as against similar models proposed in recent relevant studies. SingleLSTM is one of the simplest baseline models we use. It is made up of a single LSTM layer that is stacked above the embedding layer. A dense layer is used as classifier. A more complex model we used (BLSTM-2DCNN) was introduced in \cite{conf/coling/ZhouQZXBX16}. It makes use of a bidirectional LSTM (left and right) for feature extraction. Over that structure, 2-dimensional convolution and pooling layers are applied for feature selection. The network was exercised on various datasets and has produced top scores on sentiment analysis of short sentences. A similar and simpler model called BLSTM-POOL was proposed in \cite{Lai:2015:RCN:2886521.2886636}. It also uses a bidirectional recurrent network to extract word contexts. After that, a simple max-pooling layer serves as automatic features selector. Authors report excellent results on topic recognition task and slightly lower score on sentiment analysis of movie reviews. 
\par 
Another model we compare with, is a simple stack of few convolution layers (SingleCNN). It is very similar with the model used by Kim in \cite{DBLP:journals/corr/Kim14f}. The convolutions with filters of various lengths operate directly on word embeddings and the produced feature maps are concatenated together. Simple max-pooling and dropout layers that follow are used for feature selection and regularization. A final dense layer is used as classifier. Given that we experiment with datasets of various domains and different sizes, it would be nice to also contrast neural networks with word embedding text representations (all models above) against simple classification algorithms used over traditional bag-of-words text representation. For this reason we implemented Support Vector Machine and Logistic Regression classifier with tf-idf feature scoring, optimized with regularization parameters. 
\section{Results and Discussion}
\label{sec:resultsDiscussion}
\subsection{Sentiment Polarity Classification Results}
We ran each trial on a 70/10/20 percent split for training, development and evaluation respectively. Classification accuracy scores of all models in the four datasets are listed in Table~\ref{table:NgCnnResults}. As we can see, NgramCNN models perform relatively poor on sentiment polarity dataset which is the smallest and contains the shortest documents. Top score on that datasets (82.32\%) is achieved by BLSTM-2DCNN model presented in \cite{conf/coling/ZhouQZXBX16}. This result is slightly lower than the record of 83.1\%, reported in \cite{DBLP:journals/corr/ZhaoLP15}. Next come the optimized Logistic Regression and Support Vector Machine models, together with SingleCNN network. The worst are the models based on LSTMs only (SingleLSTM and BLSTM-POOL).
We got different results on movie reviews. Here we see that the first two NgramCNN models score over 91\%. Peak score on this dataset is 92.23\% as reported in \cite{DBLP:journals/corr/Johnson014}. We also see that Fluctuating CNN is well behind, same as the two linear models and SingleCNN. All LSTM-based models achive very low results that are below 86\%. Recurrent neural networks are obviously bad players on text analysis of long documents. 
\par 
Similar results are obtained on phone reviews, with NgramCNN basic version peaking at 95.92\%. Unfortunately, there are no relevant literature studies reporting results on this dataset. We also see that Pyramid architecture is very close, whereas Fluctuating version is again left behind. LSTM models perform relatively well, especially BLSTM-2DCNN. The simple linear models on the other hand, perform relatively bad. 
The final set of experiments is the one conducted on Yelp business reviews. Again, top result is scored from NgramCNN basic, followed by Pyramid architecture. Top literature score here is 97.36\%, reported in \cite{DBLP:conf/acl/JohnsonZ17}. It is about 2.5\% higher than that of NgramCNN Basic. SingleCNN also performed well, with accuracy of 93.86\%. The other models, including Fluctuating architecture are behind of more than 2\%.  
\begin{table}[ht] 
	\caption{Accuracy scores of NgramCNN variants and baselines}  
	\centering    
	\setlength\tabcolsep{4.7pt}  
	\begin{tabular}
		{l c c c c}  
		\toprule
		\textbf{Network} & \textbf{Sent} & \textbf{Imdb} & \textbf{Phon} & \textbf{Yelp}	\\ 
		\midrule 
		NgCNN Basic & 79.87 & 91.14 & \textbf{95.92}	& \textbf{94.95}	  \\
		NgCNN Pyramid & 79.52 & \textbf{91.21} & 95.7 & 94.83		  \\
		NgCNN Fluctuate & 77.41 & 89.32 & 93.45 & 92.27		  \\
		\hline
		Optimized LR & 81.63 & 89.48 & 92.46 & 91.75  \\
		Optimized SVM & 82.06 & 88.53 & 92.67 & 92.36  \\
		\hline
		SingleLSTM & 80.33 & 84.93 & 93.71 & 90.22	  \\ 
		SingleCNN & 81.79 & 89.84 & 94.25 & 93.86	  \\ 
		BLSTM-POOL & 80.96 & 85.54 & 94.33 & 91.19  \\
		BLSTM-2DCNN & \textbf{82.32} & 85.70 & 95.52 & 91.48  \\
		\bottomrule
	\end{tabular} 
	\label{table:NgCnnResults}
\end{table}
\subsection{Further Observations}
Considering the results of Table \ref{table:NgCnnResults}, we see that NgramCNN Basic is the best of the three designs we proposed. Pyramid version that differs only in the intermediate subsampling layers performs similarly with slightly lower accuracy scores.  
Contrary, the Fluctuating architecture performed badly on every dataset. Apparently, it is not a good neural network design. 
We also saw that models based on recurrent neural networks perform well on short documents but considerably worse on longer ones. This is probably related with their internal design. Their memory is limited and they are not able to preserve long-term word relations. 
Moreover, those kinds of networks take longer to train. CNNs that are simpler in design are also faster, with training times comparable to those of Logistic Regression and Support Vector Machine. 
These later models scored comparably well on the small datasets where they can be a good choice.
\par 
Important design choices such as network width, network depth or max-pooling region length were picked up based on the results of Section \ref{subsec:OptResults}. It is also important to note that NgramCNN basic architecture is highly flexible. Other design choices such as more convolution and pooling stacks or longer filters can be explored for better performance.  
Other network parameters were picked based on grid search results we got. Number of epochs till convergence was specific to each dataset. Yelp required nine epochs whereas phone and movie reviews converged in seven and four epochs. Finally, sentences required three epochs to converge. We did not notice much sensitivity with respect to batch size. A batch of 60 was the optimal choice in most experiments. Finally, \emph{Softplus} and \emph{Sigmoid} were equally good activation functions in the output layer.
\section{Conclusions}
\label{sec:conclusions}
In this paper, we first described the steps we followed for the creation of two datasets of song emotions that were publicly released for music emotion recognition community. We also experimented with different neural networks of convolution and max-pooling layers for analyzing sentiment polarity of different types of texts. Pretrained word embeddings of GoogleNews were utilized as generic features for a compact representation of text documents. A primary goal of our experiments was the observation of certain patterns that relate data properties to network parameters with respect to optimal classification results. The results reveal a regularity between length of documents and size of feature maps. Optimal classification scores are achieved when pooling region size is adjusted to produce feature maps that are 6 to 18 units long. Regarding datasets size, top results are achived when deeper networks are applied on the bigger datasets. Based on this result, we further examined three architecture designs we propose, comparing them with relevant baseline models. The three of them follow the design paradigm that has been highly successful in image recognition: complex feature extraction and selection combined with a simple classifier. 
Our basic version with regional max-pooling following each convolution stack resulted the most successful architecture. It is fast and flexible, easily adapting to various kinds of text data. It may thus serve as a good starting point for practitioners. A future improvement of the neural network design we proposed here, could be combining sentiment predictions of text phrases using aggregation schemes like Dempster-Shafer Inference or Abductive Reasoning \cite{citeulike:13327098}. That could be especially fruitful when analyzing sentiment polarity of long documents.   
%

%

\bibliographystyle{elsarticle-num}  
\bibliography{dta}





\end{document}